\documentclass[sigconf]{acmart}
\AtBeginDocument{%
  }

\setcopyright{acmlicensed}
\copyrightyear{2018}
\acmYear{2018}
\acmDOI{X.XX}
\acmISBN{978-1-4503-XXXX-X/2018/06}

\begin{document}

%%
%% The "title" command has an optional parameter,
%% allowing the author to define a "short title" to be used in page headers.
\title{HCMA: Hierarchical Cross-model Alignment for Grounded Text-to-Image Generation}

%%
%% The "author" command and its associated commands are used to define
%% the authors and their affiliations.
%% Of note is the shared affiliation of the first two authors, and the
%% "authornote" and "authornotemark" commands
%% used to denote shared contribution to the research.
% \author{Ben Trovato}
% \authornote{Both authors contributed equally to this research.}
% \email{trovato@corporation.com}
% \orcid{1234-5678-9012}
% \author{G.K.M. Tobin}
% \authornotemark[1]
% \email{webmaster@marysville-ohio.com}
% \affiliation{%
%   \institution{Institute for Clarity in Documentation}
%   \city{Dublin}
%   \state{Ohio}
%   \country{USA}
% }

%%
%% By default, the full list of authors will be used in the page
%% headers. Often, this list is too long, and will overlap
%% other information printed in the page headers. This command allows
%% the author to define a more concise list
%% of authors' names for this purpose.
% \renewcommand{\shortauthors}{Trovato et al.}

\author{Hang Wang}
\affiliation{%
  \institution{The Hong Kong Polytechnic University}
  \city{Hong Kong}
  \country{China}}
\email{cshwang@comp.polyu.edu.hk}

\author{Zhi-Qi Cheng}
\authornote{Corresponding author.}  % ← 新增
\affiliation{%
  \institution{University of Washington}
  \city{Seattle}
  \country{USA}}
\email{zhiqics@uw.edu}

\author{Chenhao Lin}
\affiliation{%
  \institution{Xi'an Jiaotong University}
  \city{Xi'an}
  \country{China}}
\email{linchenhao@xjtu.edu.cn}

\author{Chao Shen}
\affiliation{%
  \institution{Xi'an Jiaotong University}
  \city{Xi'an}
  \country{China}}
\email{chaoshen@mail.xjtu.edu.cn}

\author{Lei Zhang}
\affiliation{%
  \institution{The Hong Kong Polytechnic University}
  \city{Hong Kong}
  \country{China}}
\email{cslzhang@comp.polyu.edu.hk}

%%
%% The abstract is a short summary of the work to be presented in the
%% article.
\begin{abstract}
  Text-to-image synthesis has progressed to the point where models can generate visually compelling images from natural language prompts. Yet, existing methods often fail to reconcile high-level semantic fidelity with explicit spatial control, particularly in scenes involving multiple objects, nuanced relations, or complex layouts. To bridge this gap, we propose a \emph{Hierarchical Cross-Modal Alignment} (HCMA) framework for \emph{grounded} text-to-image generation. HCMA integrates two alignment modules into each diffusion sampling step: a \emph{global} module that continuously aligns latent representations with textual descriptions to ensure scene-level coherence, and a \emph{local} module that employs bounding-box layouts to anchor objects at specified locations, enabling fine-grained spatial control. Extensive experiments on the MS-COCO 2014 validation set show that HCMA surpasses state-of-the-art baselines, achieving a 0.69 improvement in Fréchet Inception Distance (FID) and a 0.0295 gain in CLIP Score. These results demonstrate HCMA’s effectiveness in faithfully capturing intricate textual semantics while adhering to user-defined spatial constraints, offering a robust solution for semantically grounded image generation. Our code is available at \url{https://github.com/hwang-cs-ime/HCMA}
\end{abstract}

%%
%% The code below is generated by the tool at http://dl.acm.org/ccs.cfm.
%% Please copy and paste the code instead of the example below.
%%
% \begin{CCSXML}
% <ccs2012>
%  <concept>
%   <concept_id>00000000.0000000.0000000</concept_id>
%   <concept_desc>Do Not Use This Code, Generate the Correct Terms for Your Paper</concept_desc>
%   <concept_significance>500</concept_significance>
%  </concept>
 % <concept>
 %  <concept_id>00000000.00000000.00000000</concept_id>
 %  <concept_desc>Do Not Use This Code, Generate the Correct Terms for Your Paper</concept_desc>
 %  <concept_significance>300</concept_significance>
 % </concept>
 % <concept>
 %  <concept_id>00000000.00000000.00000000</concept_id>
 %  <concept_desc>Do Not Use This Code, Generate the Correct Terms for Your Paper</concept_desc>
 %  <concept_significance>100</concept_significance>
 % </concept>
 % <concept>
 %  <concept_id>00000000.00000000.00000000</concept_id>
 %  <concept_desc>Do Not Use This Code, Generate the Correct Terms for Your Paper</concept_desc>
 %  <concept_significance>100</concept_significance>
 % </concept>
% </ccs2012>
% \end{CCSXML}

% \ccsdesc[300]{Do Not Use This Code~Generate the Correct Terms for Your Paper}
% \ccsdesc{Do Not Use This Code~Generate the Correct Terms for Your Paper}
% \ccsdesc[100]{Do Not Use This Code~Generate the Correct Terms for Your Paper}

%%
%% Keywords. The author(s) should pick words that accurately describe
%% the work being presented. Separate the keywords with commas.
%% A "teaser" image appears between the author and affiliation
%% information and the body of the document, and typically spans the
%% page.
% \begin{teaserfigure}
%   \includegraphics[width=\textwidth]{sampleteaser}
%   \caption{Seattle Mariners at Spring Training, 2010.}
%   \Description{Enjoying the baseball game from the third-base
%   seats. Ichiro Suzuki preparing to bat.}
%   \label{fig:teaser}
% \end{teaserfigure}

% \received{20 February 2007}
% \received[revised]{12 March 2009}
% \received[accepted]{5 June 2009}

%%
%% This command processes the author and affiliation and title
%% information and builds the first part of the formatted document.
\maketitle

\begin{figure}
  \centering
  \includegraphics[width=0.95\linewidth]{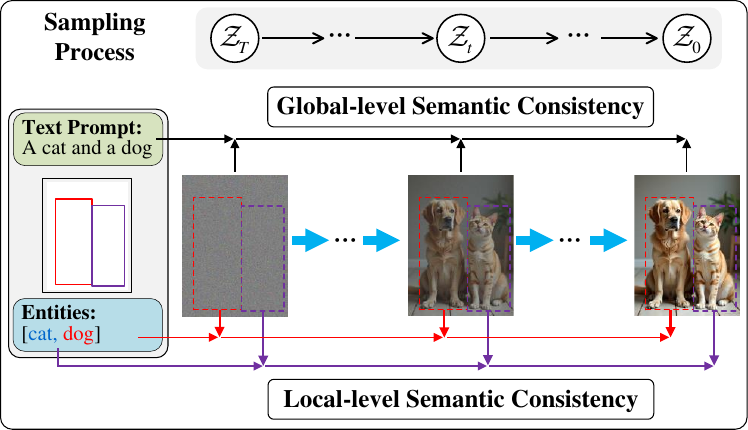}
  \vspace{-0.1in}
  \caption{\small Overview of the HCMA Approach. In each diffusion step, HCMA’s alignment ensures textual coherence globally while enforcing bounding-box layouts locally, thus providing a more robust and controllable framework for grounded text-to-image generation.}
  \label{fig:HCMA_Fig1}
  \vspace{-0.2in}
\end{figure}

\vspace{-0.1in}
\section{Introduction}
\label{sec:sec_1}
Text-to-image synthesis \cite{yuan2024generative, tan2024semantic, liao2024text, ramesh2022hierarchical, saharia2022photorealistic} has witnessed rapid advancements in recent years, enabling generative models to produce visually compelling images from natural language inputs. Nevertheless, while recent approaches often yield images that appear plausible and aesthetically appealing, they still struggle to capture the full complexity of detailed or domain-specific prompts~\cite{zhang2021cross, ye2021improving, chefer2023attend, agarwal2023star}. In particular, as scenes grow more intricate—incorporating multiple objects, nuanced relationships, or fine-grained constraints—existing solutions frequently falter, underscoring the need for more powerful and controllable text-to-image frameworks.

A promising strategy for enhancing control in text-to-image generation involves specifying additional spatial constraints, especially bounding boxes that delineate where objects should appear. This approach underlies \emph{grounded text-to-image generation}, wherein the model must simultaneously satisfy global textual semantics and explicit geometric layouts. Such grounded methods have broad utility, ranging from interactive design and storytelling \cite{carstensdottir2020progression, li2022interactive, liu2024intelligent, wang2024evolving, zeng2024peeling, singh2025storybooth} to augmented and virtual reality \cite{li2023location, boroushaki2023augmenting, ai2024characterizing}, in which precise object placement is essential for user immersion or functional coherence.

Over the past few years, various grounded text-to-image frameworks have emerged~\cite{phung2024grounded, li2023gligen, xie2023boxdiff, chen2024training, zhao2023loco}, often relying on specialized cross-attention mechanisms, gating modules, or other custom constraints to ensure bounding-box compliance. For example, GLIGEN~\cite{li2023gligen} integrates a gated self-attention mechanism to enhance spatial controllability, and BoxDiff~\cite{xie2023boxdiff} utilizes inner- and outer-box cross-attention constraints to adhere to user-defined object layouts. Alternatively, Attention-Refocusing~\cite{phung2024grounded} harnesses GPT-4~\cite{openai2023gpt4} to propose object placements before refining cross-attention and self-attention layers. Despite these innovations, most existing methods do not guarantee both robust \emph{global} text-image coherence and \emph{local} bounding-box fidelity at every stage of the generation process. Consequently, while a system may capture the overall scene described by the text, individual regions often deviate from user-prescribed positional or categorical requirements.

To address these limitations, we present a \emph{Hierarchical Cross-Modal Alignment} (HCMA) framework that enforces semantic consistency at both the global and local scales during grounded text-to-image generation. As detailed in {Section~\ref{sec:sec_3} and illustrated in Figure~\ref{fig:HCMA_Fig1}, HCMA weaves two complementary alignment modules into each step of a diffusion-based sampling procedure. 
\textbf{(1)~Global-Level (Caption-to-Image) Alignment} continuously matches the evolving latent representation with the textual prompt, thereby preserving high-level semantic coherence across the entire scene. Rather than verifying alignment only at the final step, it iteratively refines the latent space to reflect overarching themes and relationships in the text, benefiting simple captions and elaborate, multi-object prompts alike. 
\textbf{(2)~Local-Level (Object-to-Region) Alignment} operates in parallel, leveraging bounding-box layouts to anchor each object at its designated location. At each iteration, it ensures localized latent features adhere to user-defined constraints, preventing objects from drifting away from prescribed positions or diverging from specified attributes. This layered mechanism proves especially crucial for prompts requiring strict spatial arrangements, such as non-overlapping objects or distinct, non-negotiable placements for certain categories.

By merging these two alignment processes throughout the diffusion steps, HCMA delivers both \emph{semantic rigor} at the scene level and \emph{precise controllability} at the object level. The global module maintains coherence with the entire textual narrative, while the local module enforces bounding-box constraints to uphold fine-grained spatial accuracy. This layered strategy effectively tackles the dual challenge of capturing a prompt’s high-level semantics and adhering to region-level instructions—requirements frequently encountered in real-world tasks like interactive storyboarding, VR world-building, or industrial design layout.

\noindent \textbf{Our main contributions are as follows:}
\begin{itemize}
    \item \textbf{HCMA Framework.} We propose \textbf{HCMA}, a grounded text-to-image generation framework centered on \emph{hierarchical cross-modal alignment}, ensuring both semantic fidelity and spatial controllability. By disentangling global and local alignment processes, HCMA effectively preserves textual semantics while adhering to bounding-box constraints.
    \item \textbf{Dual-Alignment Strategy.} We introduce a two-tiered alignment mechanism, comprising a global-level \emph{caption-to-image} module and a local-level \emph{object-to-region} module. Each component is optimized with distinct losses to maintain overall prompt coherence and strict, precise object placement throughout the diffusion sampling.
    \item \textbf{Extensive Experimental Validation.} Across the MS-COCO 2014 validation set, HCMA surpasses leading baselines, achieving a 0.69 improvement in FID and a 0.0295 gain in CLIP Score. These results confirm the framework’s effectiveness in producing images that remain coherent at the macro level while respecting bounding-box constraints at the micro level.
\end{itemize}

\section{Related Work}
\label{sec:related_work}

\subsection{Text-to-Image Generation}
Text-to-image generation aims to synthesize images that accurately capture the semantic content of a given textual prompt~\cite{li2019controllable, zhang2023iti, liu2023detector, zhou2023shifted}. Early approaches often faced challenges in scaling to high-resolution outputs without sacrificing semantic fidelity.  
A major breakthrough came with the Latent Diffusion Model (LDM)~\cite{rombach2022high}, which projects the generation process from high-dimensional pixel space to a lower-dimensional latent space, substantially reducing computational overhead.  
Building on this idea, Stable Diffusion Models v1.4 and v1.5 introduced optimized sampling strategies and improved pre-trained VAEs, delivering higher-fidelity images and greater user control. Subsequent versions, such as v2, v3, and SD-XL, further refined the architecture with advanced attention mechanisms and higher-resolution capabilities, enabling more detailed and context-aware synthesis for complex prompts.

Beyond these diffusion-based models, LAFITE~\cite{zhou2022towards} demonstrated versatility in zero-shot and language-free scenarios by leveraging a sophisticated architecture that spans multiple semantic domains. 
Gafni et al.~\cite{gafni2022make} incorporated human priors to improve generation quality, emphasizing salient regions and leveraging perception-level insights. 
Additionally, Kang et al.~\cite{kang2023counting} tackled the persistent issue of object miscounting by introducing an attention map-based guidance approach, enabling high-fidelity outputs with accurate object counts dictated by the prompt. 
Despite these advancements, most existing methods primarily focus on global text-image coherence, often overlooking the need for precise semantic alignment at finer spatial or object-centric levels.

\begin{figure*}
  \centering
  \includegraphics[width=0.95\textwidth]{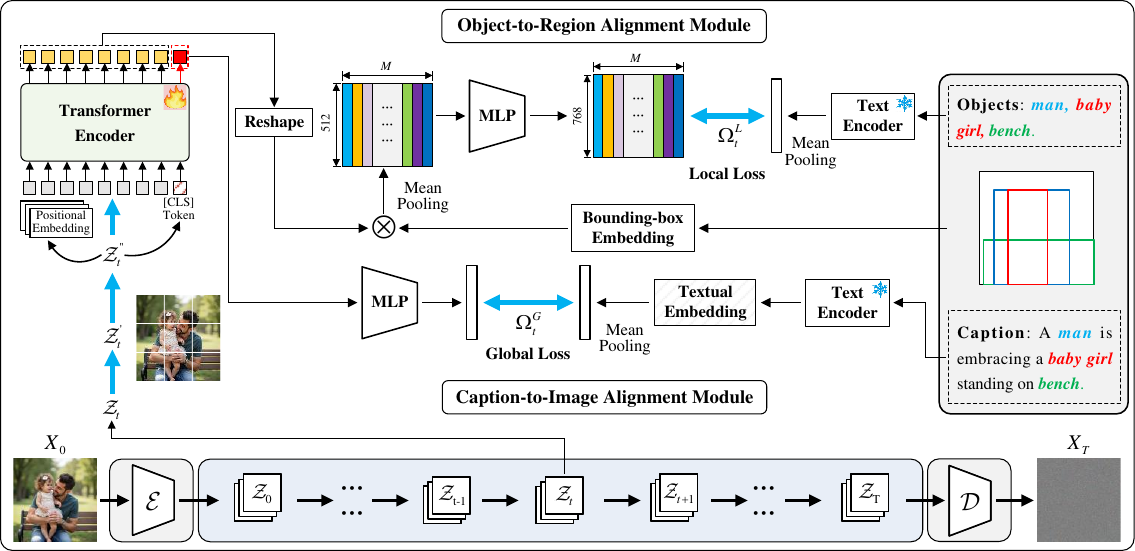}
  \vspace{-0.1in}
  \caption{\small Training Framework of HCMA. Given a text prompt and bounding-box layouts, a \emph{global} (caption-to-image) module fosters textual coherence while a \emph{local} (object-to-region) module enforces bounding-box adherence. They guide the diffusion-based training so each step reflects both high-level semantics and region-specific placement.}
  \vspace{-0.1in}
  \label{fig:HCMA_Framework_Training}
\end{figure*}

\subsection{Grounded Text-to-Image Generation}
Grounded text-to-image generation~\cite{zhao2023loco, chen2024training, bar2023multidiffusion, sun2024spatial, phung2024grounded} extends the basic text-to-image paradigm by specifying object placements or spatial layouts alongside textual descriptions. This task requires the synthesized images to maintain overall semantic faithfulness while accurately arranging objects according to predefined bounding boxes or other geometric constraints. 
LoCo~\cite{zhao2023loco} used semantic priors embedded in padding tokens to reinforce alignment between image generation and input constraints. 
Layout-Guidance~\cite{chen2024training} introduced a training-free framework that modifies cross-attention layers for both forward and backward guidance, leading to fine-grained control over layout. Meanwhile, MultiDiffusion~\cite{bar2023multidiffusion} formulated a numerical optimization problem based on a pre-trained diffusion model to enhance controllable image quality. Similarly, SALT-AG~\cite{sun2024spatial} replaced random noise with spatially aware noise and adopted attention-guided regularization to improve layout adherence in synthesized outputs.

Although these methods achieve varying degrees of spatial controllability, most do not explicitly enforce both global text-image coherence and local object-level alignment. Our work focuses on establishing hierarchical semantic consistency at both the image and region levels, thereby achieving more robust cross-modal alignment under complex textual and spatial constraints.

\section{Methodology}
\label{sec:sec_3}
In this section, we introduce our proposed \emph{Hierarchical Cross-Modal Alignment (HCMA)} framework for grounded text-to-image generation. We begin with an overview of diffusion-based image synthesis approaches and the problem setup (Section~\ref{sec:preliminaries}--\ref{subsec:problem_formulation}), followed by a detailed description of the HCMA model architecture and its training/inference procedures (Section~\ref{subsec:hcma_model}).

\begin{figure*}
  \centering
  \includegraphics[width=0.95\textwidth]{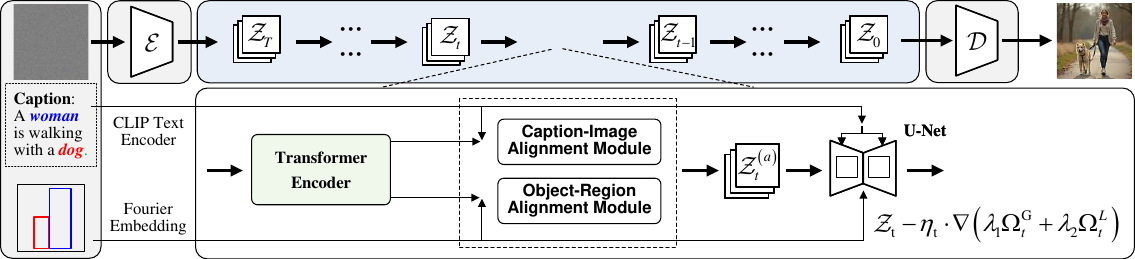}
  \vspace{-0.1in}
  \caption{%
  % \textbf{Sampling Process of HCMA.}
  \small Sampling Process of HCMA. During inference, each diffusion step systematically alternates between refining global semantics via the caption-to-image module and strictly enforcing local bounding-box fidelity via the object-to-region module. The final image thus comprehensively unifies the textual description with explicitly defined spatial constraints.}
  \vspace{-0.1in}
  \label{fig:HCMA_Framework_Testing}
\end{figure*}

\subsection{Preliminaries}
\label{sec:preliminaries}
Diffusion-based methods have recently emerged as a powerful paradigm in text-to-image generation. Among these, the Latent Diffusion Model~\cite{rombach2022high} (LDM) introduced the idea of operating in a learned latent space rather than raw pixel space, enabling more efficient and scalable image synthesis under various conditioning inputs (e.g., text, reference images, spatial layouts).
Building on LDM, \emph{Stable Diffusion} leverages a subset of the LAION-5B dataset~\cite{schuhmann2022laion} and employs a CLIP text encoder~\cite{radford2021learning}, thereby effectively bridging text and image features and substantially improving the interpretation of complex prompts and the realism of synthesized images. Formally, given a latent representation $\mathbf{Z}_t$ at time step $t$, the stable diffusion training objective aims to learn a noise estimation function $\epsilon_\theta$ that predicts the Gaussian noise $\epsilon$ added at each diffusion step:
\begin{equation}
   \mathop{\arg\min}\limits_{\theta} \| \epsilon - \epsilon_{\theta} \left( \mathbf{{Z}_t}, t,  f^T(c) \right) \| ^2,
\end{equation}
where $f^T(c)$ is the CLIP embedding of $c$. At inference, a U-Net~\cite{ronneberger2015u} with a spatial transformer fuses the latent representation $\mathbf{Z}_t$ and the text embedding $\mathbf{c}$ via cross-attention:
\begin{align}
   &\mathbf{Q} = \mathbf{Z}_t \cdot \mathbf{W}^Q, \mathbf{K} = \mathbf{c} \cdot \mathbf{W}^K, \mathbf{V} = \mathbf{c} \cdot \mathbf{W}^V, \\
   &\text{Attention}(\mathbf{Z}_t, \mathbf{c}) =  \text{Softmax}\left( \frac{\mathbf{Q}\mathbf{K}^T}{\sqrt{d}} \right) \cdot \mathbf{V},
\end{align}
where $\mathbf{W}^Q$, $\mathbf{W}^K$, and $\mathbf{W}^V$ are learnable projection matrices, and $d$ is the channel dimension. Repeated cross-attention steps integrate textual context into the latent image features, guiding the denoising process to generate text-aligned, high-quality images.

\subsection{Problem Formulation}
\label{subsec:problem_formulation}
We focus on \emph{grounded text-to-image generation}, which aims to synthesize images that align with both textual descriptions and bounding-box constraints. Consider a noisy image $X_T$, a text description $c = \{c_1, c_2, \dots, c_n\}$, and a set of bounding boxes $\mathcal{B} = \{b_1, b_2, \dots, b_M\}$, where $M$ is the number of bounding boxes. Each bounding box specifies the location of an object, whose category label $y_i$ is encoded via CLIP to produce $f^T(y_i)$, where $y = \{y_1, \ldots, y_M\}$. A pre-trained VAE transforms the noisy image $X_T$ into a latent representation $\mathbf{Z}_T$, and each box $b_i$ is mapped to $\mathbf{B}$ using Fourier embeddings~\cite{nerf}.

In contrast to previous diffusion-based techniques that typically fuse text guidance and denoising in a single pass, our method partitions each diffusion step $t$ into two complementary sub-processes, thereby ensuring more precise alignment with both global text semantics and local object constraints. Specifically:
\begin{enumerate}
    \item \textbf{Hierarchical semantic alignment}, which aligns the latent representation with both the text prompt and each bounding box’s object category. This ensures that the refined latent $\mathbf{Z}_t^{(a)}$ respects global (image-level) semantics and local (region-level) constraints.
    \item \textbf{Standard denoising}, in which the U-Net robustly predicts and removes the noise at step $t$, eventually producing $\mathbf{Z}_{t-1}$ from the newly aligned latent $\mathbf{Z}_t^{(a)}$.
\end{enumerate}

Mathematically, these two processes can be expressed as:
\begin{numcases}{}
    \mathbf{Z}_t^{(a)} = \mathcal{H}\bigl(\mathbf{Z}_t, t, c, \mathcal{B}, y\bigr), \label{eq:alignment} \\
    \mathbf{Z}_{t-1} = \mathcal{G}\bigl(\mathbf{Z}_t^{(a)}, t, f^T(c), \mathbf{B}, f^T(y), \epsilon_{\theta} \bigr), \label{eq:denoising}
\end{numcases}
where $\mathbf{Z}_t$ is the noisy latent at step $t$, $\mathbf{Z}_t^{(a)}$ is the aligned latent after applying the alignment module $\mathcal{H}$, and $\mathcal{G}$ denotes the U-Net’s denoising process. By iterating these alternating sub-processes, the model progressively refines the latent representation, ensuring that the final output respects both global textual semantics and local bounding-box specifications. This iterative refinement yields a high-quality image that remains faithful to the provided grounding conditions at each level of granularity.

\subsection{HCMA Model}
\label{subsec:hcma_model}
In this section, we introduce our \emph{Hierarchical Cross-Modal Alignment (HCMA)} model. HCMA interleaves dedicated \emph{alignment} and \emph{denoising} steps during diffusion sampling to address both \emph{global semantics} (caption-to-image) and \emph{local layouts} (object-to-region). Through this layered alignment strategy, HCMA ensures the synthesized image preserves overarching textual consistency while affording precise spatial control over individual objects.

\noindent \textbf{Overall Architecture.}
As illustrated in Figure~\ref{fig:HCMA_Framework_Training}, consider an input image $X \in \mathbb{R}^{3 \times H \times W}$, a text description $c \in \mathbb{R}^n$, and a set of bounding boxes $\mathcal{B} \in \mathbb{R}^{M \times 4}$. We extract a latent variable $\mathbf{Z}_t \in \mathbb{R}^{4 \times \frac{H}{8} \times \frac{W}{8}}$ from a randomly chosen diffusion step $t$. Each bounding box $b_i$ is transformed via a Fourier Embedding into $\mathbf{B} \in \mathbb{R}^{\frac{H}{32} \times \frac{W}{32} \times M}$. A Vision Transformer (ViT) then encodes $\mathbf{Z}_t$, yielding rich multi-scale visual features, which are subsequently passed through our hierarchical alignment module consisting of:
\begin{enumerate}
    \item \textbf{Caption-Image Alignment (C2IA)}: Enforces semantic coherence at a global level by aligning the latent image representation with the textual prompt.
    \item \textbf{Object-Region Alignment (O2RA)}: Enforces spatial and categorical consistency at a local level, aligning bounding-box regions with their respective object classes.
\end{enumerate}
By explicitly injecting both global and local alignment signals into the diffusion sampling procedure, HCMA achieves a balanced trade-off between holistic text-image fidelity and fine-grained control over specific visual objects.

\noindent \textbf{Latent Feature Extraction.}
At diffusion step $t$, the latent $\mathbf{Z}_t$ is segmented and projected into $\mathbf{Z}_t' \in \mathbb{R}^{N \times d_k}$, where $N = \tfrac{H}{32} \times \tfrac{W}{32}$ and $d_k = 64$. An MLP then transforms $\mathbf{Z}_t'$ into a set of visual tokens $\mathbf{Z}_t''$. We prepend a learnable [CLS] token to these tokens, add positional embeddings, and feed the resulting sequence into the ViT. The ViT output naturally decomposes into two parts: a global representation $\mathbf{Z}_t^G \in \mathbb{R}^{d_v}$, associated with the [CLS] token and capturing high-level scene semantics, and a local representation $\mathbf{Z}_t^L \in \mathbb{R}^{N \times d_v}$, containing patch-level details for subsequent object-region alignment. \emph{Both global and local representations serve as the basis for the hierarchical alignment mechanisms described next, ensuring the model effectively incorporates textual semantics and object-level constraints.}

\subsubsection{Caption-to-Image Alignment \& Global-Level Loss}
Achieving overall fidelity between the synthesized image and the textual description remains a key and persistent challenge in text-to-image generation. In the absence of a robust global constraint, a model may accurately align certain local regions to segments of the text while still failing to capture the broader thematic or contextual elements of the prompt. The \emph{Caption-to-Image Alignment (C2IA)} module addresses this shortcoming by projecting both the latent image representation and the text embedding into a shared semantic space, thereby effectively promoting holistic coherence.

Concretely, we obtain $\mathbf{Z}_t^G$ (the global representation) from the ViT’s [CLS] token and project it through a three-layer fully connected network, producing $\mathbf{f}_t^G \in \mathbb{R}^{d_t}$. In parallel, the text prompt $c$ is encoded by the CLIP text encoder, yielding $\mathbf{f}^T(c)$. To quantify global alignment, we define:
\begin{equation}
\Omega_t^G = 1 - \frac{\mathbf{f}_t^G \cdot f^T(c)}{\|\mathbf{f}_t^G\|\;\|f^T(c)\|}.
\end{equation}
Here, $\Omega_t^G$ effectively measures the inverse cosine similarity between $\mathbf{f}_t^G$ and $f^T(c)$. By minimizing $\Omega_t^G$, the model incrementally refines the latent image’s global features to more closely match the text embedding. Consequently, each diffusion step is guided toward producing images that reflect not only local details but also the overarching themes and context described by $c$.

\subsubsection{Object-to-Region Alignment \& Local-Level Loss}
Even with robust global alignment, misplacement or misclassification of objects can occur if local constraints are overlooked. \emph{Object-to-Region Alignment (O2RA)} addresses this by tying specific bounding-box regions to their corresponding object categories, preserving local fidelity to the textual instructions.

We reshape $\mathbf{Z}_t^L$ into $\mathbf{Z}_t^l \in \mathbb{R}^{\frac{H}{32}\times \frac{W}{32}\times d_v}$ and fuse it with the bounding-box feature $\mathbf{B}\in \mathbb{R}^{\frac{H}{32}\times \frac{W}{32}\times M}$. This produces $\mathbf{Z}_t^{L_b} \in \mathbb{R}^{\frac{H}{32}\times \frac{W}{32}\times M\times d_v}$. After mean pooling and an MLP projection, we obtain $\mathbf{f}_t^L \in \mathbb{R}^{M\times d_t}$, where $\mathbf{f}_{t,b_i}^L$ corresponds to the $i$-th bounding box. If $f^T(y_i)$ is the CLIP embedding of the label $y_i$, the \emph{local-level loss} is:
\begin{equation}
\Omega_t^L = \frac{1}{M}\,\sum_{i=1}^M \left( 1 - \frac{\mathbf{f}_{t,b_i}^L \cdot f^T(y_i)}{\|\mathbf{f}_{t,b_i}^L\|\;\|f^T(y_i)\|} \right).
\end{equation}
Consequently, minimizing $\Omega_t^L$ drives the model to generate objects in each bounding-box region that accurately match their designated categories and spatial extents.

\subsubsection{Sampling Update in Latent Space}
During inference, HCMA interleaves \emph{alignment} and \emph{denoising} at each diffusion step $t$. Let $\mathbf{Z}_t$ denote the latent representation at step $t$. We first apply an \emph{alignment} update:
\begin{equation}
\mathbf{Z}_t^{(a)} \leftarrow \mathbf{Z}_t -
\nabla \Bigl(\lambda_1\,\Omega_t^G + \lambda_2\,\Omega_t^L\Bigr)\,\eta_t,
\end{equation}
where $\lambda_1$ and $\lambda_2$ balance the global and local losses, and $\eta_t$ is the alignment learning rate. This step refines $\mathbf{Z}_t$ to better satisfy both caption-to-image and object-to-region constraints. Next, we pass the updated latent $\mathbf{Z}_t^{(a)}$ to a U-Net $\mathcal{G}$ for a \emph{denoising} update:
\begin{equation}
\mathbf{Z}_{t-1} \leftarrow \mathbf{Z}_t^{(a)} - \gamma\,\epsilon_\theta\Bigl(\mathbf{Z}_t^{(a)},\, t,\, f^T(c),\, \mathbf{B}\Bigr),
\end{equation}
where $\gamma$ is a step-size parameter, and $\epsilon_\theta$ is the learned noise-estimation model. By iterating from $t = T$ down to $t = 0$, we perform a ``dual refinement'' at every step---alternating between alignment and denoising---to ultimately obtain the final latent $\mathbf{Z}_0$. This refined latent is then decoded by the VAE into the synthesized image, ensuring that the output respects the textual semantics and bounding-box constraints while maintaining high visual fidelity.

% \newpage
\begin{algorithm}[!tb]
\caption{Hierarchical Cross-modal Alignment (HCMA) Training and Inference}
\label{alg:hcma}
\begin{algorithmic}[1]
\State \textbf{Input:} Text prompt $c$, bounding boxes $B = \{b_i\}_{i=1}^M$, object categories $y = \{y_i\}_{i=1}^M$, total denoising steps $T$, initial latent $\mathbf{Z}_T$, learning rate $\eta_t$
\State \textbf{Output:} Generated image latent $\mathbf{Z}_0$

\State \textbf{Training:}
\For{each training step $t$}
    \State Encode text prompt $c$ using CLIP text encoder: $\mathbf{f}_T(c)$
    \State Encode object categories $y_i$ using CLIP: $\mathbf{f}_T(y_i)$
    \State Compute bounding box embedding $B$ with Fourier encoding
    \State Extract visual tokens from $\mathbf{Z}_t$ via ViT to get $\mathbf{Z}^G_t$, $\mathbf{Z}^L_t$
    \State Project $\mathbf{Z}^G_t$ and $\mathbf{Z}^L_t$ through MLPs to get $\mathbf{f}^G_t$, $\mathbf{f}^L_t$
    \State Compute global alignment loss: $\Omega^G_t = 1 - \cos(\mathbf{f}^G_t, \mathbf{f}_T(c))$
    \State Compute local alignment loss: $\Omega^L_t = \frac{1}{M} \sum_{i=1}^M \left(1 - \cos(\mathbf{f}^L_{t,b_i}, \mathbf{f}_T(y_i))\right)$
    \State Compute alignment-guided latent: $\mathbf{Z}_t^{(a)} \gets \mathbf{Z}_t - \eta_t \cdot \nabla (\lambda_1 \Omega^G_t + \lambda_2 \Omega^L_t)$
    \State Update noise prediction model $\epsilon_\theta$ using noise loss $\|\epsilon - \epsilon_\theta(\mathbf{Z}_t^{(a)}, t, \mathbf{f}_T(c), B)\|^2$
\EndFor

\State \textbf{Inference (Sampling):}
\State Initialize latent $\mathbf{Z}_T \sim \mathcal{N}(0, I)$
\For{$t = T$ \textbf{to} $1$}
    \State Compute $\Omega^G_t$, $\Omega^L_t$ as in training
    \State Compute aligned latent: $\mathbf{Z}_t^{(a)} \gets \mathbf{Z}_t - \eta_t \cdot \nabla (\lambda_1 \Omega^G_t + \lambda_2 \Omega^L_t)$
    \State Predict noise: $\epsilon_\theta(\mathbf{Z}_t^{(a)}, t, \mathbf{f}_T(c), B)$
    \State Update latent: $\mathbf{Z}_{t-1} \gets \mathbf{Z}_t^{(a)} - \gamma \cdot \epsilon_\theta(\cdot)$
\EndFor
\State \Return{$\mathbf{Z}_0$}
\end{algorithmic}
\end{algorithm}

\subsection{Training and Inference}
Algorithm~\ref{alg:hcma} outlines the overall procedure. 
\textbf{(1)}~During training, at each diffusion step \(t\), the model computes the global and local alignment losses (\(\Omega_t^G\) and \(\Omega_t^L\)), updates \(\mathbf{Z}_t\) accordingly, and then applies the standard diffusion noise-prediction objective to optimize \(\epsilon_\theta\). After sufficient epochs, HCMA becomes adept at integrating textual semantics and bounding-box constraints within the latent space. 
\textbf{(2)}~At inference, \(\mathbf{Z}_T\) is randomly initialized and subjected to an \emph{align-then-denoise} routine, iterating until it converges to a latent embedding that satisfies both the global textual prompt and the specified bounding boxes. This iterative process yields a final latent state \(\mathbf{Z}_0\), which the VAE decodes into an image exhibiting semantic fidelity to the text and accurately placing objects as prescribed. 
\textbf{(3)}~By integrating hierarchical cross-modal alignment into a diffusion-based sampling framework, HCMA offers a flexible and robust solution for generating images under complex textual descriptions and bounding-box layouts. It enforces caption-to-image and object-to-region constraints, mitigating issues such as object misclassification or mismatched scenes. Furthermore, by alternating alignment and denoising steps, the model incrementally incorporates textual and layout guidance, enabling reliable multi-object or multi-layout generation without sacrificing image quality.

\section{Experiments}
\label{sec:sec_4}

\begin{table*}[htbp]
  \centering
  \begin{tabular}{l | c c | c c}
    \toprule
                     \multirow{2}{*}{Methods}       &\multicolumn{2}{c}{Image-level}                                 &\multicolumn{2}{c}{Region-level}                                      \\
                                                    \cmidrule(r){2-3}                                                \cmidrule(l){4-5}
                                                    & FID ($\downarrow$)           & CLIP score ($\uparrow$)         & FID ($\downarrow$)                 & CLIP score ($\uparrow$)         \\
    \midrule
    SD-v1.4~\cite{rombach2022high}                  & 12.12                        & 0.2917                          & 13.66                              & 0.2238                          \\
    SD-v1.5~\cite{rombach2022high}                  & 9.43                         & 0.2907                          & 13.76                              & 0.2237                          \\
    Attend-and-Excite~\cite{chefer2023attend}       & 14.87                        & 0.2854                          & 14.57                              & 0.2224                          \\
    BoxDiff~\cite{xie2023boxdiff}                   & 20.26                        & 0.2833                          & 14.27                              & 0.2248                          \\
    Layout-Guidance~\cite{chen2024training}         & 21.82                        & 0.2869                          & 21.50                              & 0.2262                          \\
    GLIGEN (LDM)~\cite{li2023gligen}                & 13.05                        & 0.2891                          & 18.53                              & 0.2381                          \\
    GLIGEN (SD-v1.4)~\cite{li2023gligen}            & 12.07                        & 0.2936                          & 17.25                              & 0.2356                          \\
    \hline \hline
    \textbf{HCMA}(SD-v1.4)                          & \textbf{8.74}                & \textbf{0.3231}                 & \textbf{12.48}                     & \textbf{0.2392}                 \\

  \bottomrule
\end{tabular}
\caption{%
\small Quantitative comparison on the MS-COCO 2014 validation set, evaluated at both image-level and region-level.
“SD-v1.4” and “SD-v1.5” denote backbone diffusion models v1.4 and v1.5, while “LDM” refers to the original latent diffusion model \cite{rombach2022high}.
Lower FID ($\downarrow$) and higher CLIP score ($\uparrow$) indicate better performance, with bolded figures indicating best metrics.
}
\vspace{-2.0em} % 根据需要调整负值
\label{tab:baselines}
\end{table*}

\subsection{Datasets}
We employ the MS-COCO 2014 dataset \cite{lin2014microsoft} for our experiments, comprising 82{,}783 images for training and 40{,}504 images for validation. Each image features five textual captions describing its semantic content, as well as bounding boxes and category labels for the objects present. This extensive annotation scheme offers a robust platform for evaluating both global-level semantic alignment and local-level object consistency in grounded text-to-image generation.

For our HCMA approach, we partition the MS-COCO 2014 training set into 80\% for training and 20\% for validation, maintaining consistency with standard practices in text-to-image research. Following typical protocols for grounded text-to-image generation, we then use the official MS-COCO 2014 validation set as our testing set. To enable direct comparison with prior methods and to ensure a comprehensive assessment of our framework, we sample 30{,}000 images from the test set to report final quantitative results. 

In all experiments, we adopt the Fréchet Inception Distance (FID)~\cite{heusel2017gans} and CLIP score~\cite{radford2021learning} to measure performance. FID quantifies the distributional similarity between real and synthesized images, while the CLIP score evaluates semantic consistency by comparing text and image embeddings within a shared feature space. Together, these metrics provide an in-depth analysis of both visual fidelity and text alignment.

\begin{table*}[htbp]
  \centering
  \begin{tabular}{c c | c c c c c c}
    \toprule
    \multirow{2}{*}{C2IA}          & \multirow{2}{*}{O2RA}      &\multicolumn{2}{c}{Image-level}                          &\multicolumn{2}{c}{Region-level}                           \\
                                                                \cmidrule(r){3-4}                                         \cmidrule(l){5-6}
                                   &                            & FID ($\downarrow$)    & CLIP score ($\uparrow$)         & FID ($\downarrow$)           & CLIP score ($\uparrow$)    \\
    \midrule
                  &                                             & 12.07                 & 0.2936                          & 17.25                        & 0.2356                     \\
                  & \Checkmark                                  & 9.62                  & 0.3225                          & 11.65                        & 0.2389                     \\
    \Checkmark    &                                             & 9.14                  & 0.3229                          & 11.75                        & 0.2386                     \\
    \Checkmark    & \Checkmark                                  & \textbf{8.74}         & \textbf{0.3231}                 & \textbf{12.48}               & \textbf{0.2392}            \\

  \bottomrule
\end{tabular}
\caption{%
\small Ablation study on the MS-COCO 2014 validation set, showing how each alignment module (\emph{C2IA} for global-level, \emph{O2RA} for local-level) affects image- and region-level performance. Combining both modules yields the best FID and CLIP metrics.
}
\vspace{-2.0em} % 根据需要调整负值
\label{tab:ablation}
\end{table*}

\subsection{Evaluation Metrics and Baselines}

\noindent \textbf{Fréchet Inception Distance (FID)}~\cite{heusel2017gans}: 
We use FID to quantify how closely generated outputs resemble real images at a distributional level, using feature embeddings from Inception-V3. Specifically, FID computes the Fréchet distance between the mean and covariance of embeddings from the real and synthesized image sets. Lower FID values indicate greater similarity to real images, reflecting higher visual fidelity and diversity.

\noindent \textbf{CLIP Score}~\cite{radford2021learning}:
We also evaluate semantic alignment between generated images and text prompts through the CLIP model (Contrastive Language-Image Pre-training). By embedding both images and text in a shared feature space, CLIP computes the cosine similarity between generated images and their corresponding textual descriptions. Higher CLIP scores thus indicate stronger semantic consistency. For each compared method, we use a pre-trained CLIP ViT-B/32 model to ensure consistent evaluation.

\noindent \emph{\textbf{Baseline Methods}.}
We compare HCMA against seven state-of-the-art models, following their official training and validation setups on the MS-COCO 2014 dataset and evaluating on its official validation set. Table~\ref{tab:baselines} presents the final results. Specifically:
% \vspace{-0.1in}
\begin{itemize}[leftmargin=*, itemsep=0pt]
  \item \textbf{SD-v1.4}~\cite{rombach2022high}, \textbf{SD-v1.5}~\cite{rombach2022high}, and \textbf{Attend-and-Excite}~\cite{chefer2023attend} are text-to-image generation methods relying solely on textual prompts.
  \item \textbf{BoxDiff}~\cite{xie2023boxdiff}, \textbf{Layout-Guidance}~\cite{chen2024training}, \textbf{GLIGEN (LDM)}~\cite{li2023gligen}, and \textbf{GLIGEN (SD-v1.4)}~\cite{li2023gligen} are grounded text-to-image approaches that accept bounding box layouts in addition to the text prompt.
\end{itemize}
% \vspace{-0.1in}
We adopt each method’s official hyperparameters to ensure fairness and reproducibility. This unified evaluation strategy allows us to rigorously assess HCMA’s performance with respect to both semantic fidelity and spatial adherence.

\subsection{Implementation Details}
We build HCMA upon Stable Diffusion v1.4 as our primary diffusion backbone. During training, we employ a batch size of 1024 on an HPC environment, allowing efficient parallelization. Our Vision Transformer (ViT) model consists of 8 attention heads, 6 transformer blocks, a hidden dimension of $d_v=512$, and an input resolution of $64\times64$. Training proceeds for 60 epochs using the Adam optimizer (learning rate $1\times10^{-3}$, weight decay $5\times10^{-4}$), and the hyperparameters $\lambda_1$ and $\lambda_2$ are set to 1. 

Regarding the sampling process, we utilize the PLMS sampler with 50 steps. At each diffusion step, we alternate between alignment optimization and denoising, thereby ensuring that each partial update thoroughly preserves semantic integrity. This iterative procedure consistently promotes alignment across both global and local levels of the generated images. Throughout experimentation, we maintain fixed random seeds to reinforce reproducibility and systematically monitor performance at regular intervals to select stable checkpoints. Our final model thus balances alignment accuracy and computational efficiency while preserving high visual fidelity across diverse scenes and object arrangements.

\subsection{Quantitative Results}
Table~\ref{tab:baselines} compares our HCMA framework with seven state-of-the-art baselines. Specifically, SD-v1.4, SD-v1.5, and Attend-and-Excite are text-to-image generation methods, whereas BoxDiff, Layout-Guidance, GLIGEN (LDM), and GLIGEN (SD-v1.4) address grounded text-to-image tasks. From these results, HCMA attains the highest performance in both FID and CLIP scores, indicating its effectiveness at both the image and region levels.

\noindent \textbf{Image-Level Performance.}~Compared to SD-v1.5, HCMA improves FID by 0.69 and CLIP score by 0.0324, demonstrating enhanced semantic consistency and higher visual fidelity across generated scenes. Against the strongest grounded method, GLIGEN (SD-v1.4), HCMA achieves gains of 3.33 in FID and 0.0295 in CLIP score. These substantial margins reflect HCMA’s capability to handle finer object interactions, complex scene composition, and detailed text instructions that other methods often struggle with. The improvements likely stem from the synergy between global semantic alignment and local bounding-box adherence, where each diffusion step benefits from both high-level coherence and precise spatial constraints.

\noindent \textbf{Region-Level Performance.}~Focusing on local quality, HCMA surpasses SD-v1.5 by 1.28 in FID and 0.0155 in CLIP score, underscoring its prowess in preserving object integrity and aligning local semantics. Compared to GLIGEN (SD-v1.4), HCMA yields additional gains of 4.77 in FID and 0.0036 in CLIP score, highlighting the efficacy of hierarchical cross-modal alignment in enforcing strict spatial constraints without sacrificing realism. Overall, these findings confirm that unifying global (image-wide) and local (region-level) alignment enables HCMA to faithfully capture multi-object prompts and fulfill complex layout requirements in a cohesive manner.

\subsection{Ablation Study}
To investigate the roles of the caption-to-image alignment (C2IA) and object-to-region alignment (O2RA) modules in our proposed HCMA framework, we conduct ablation experiments by systematically disabling one or both modules. The key findings, presented in Table~\ref{tab:ablation}, reveal how each module contributes to both image-level and region-level performance.

\subsubsection{Effectiveness of the C2IA Module}
Removing C2IA focuses the entire alignment on local bounding-box adherence, leaving O2RA as the sole mechanism for enforcing semantic consistency. Consequently, the variant HCMA (w/o C2IA) aligns object categories and bounding boxes at each diffusion step, but lacks an explicit strategy to ensure global semantic coherence between the generated image and the text prompt. 
From Table~\ref{tab:ablation}, we observe that without C2IA, the FID score degrades by 0.88 and the CLIP score decreases by 0.0006 at the image level; region-level CLIP also drops by 0.0003. Despite this decline, HCMA (w/o C2IA) still achieves considerable gains over the approach missing both modules, indicating that even local alignment alone can provide meaningful improvements. 
However, these results underscore the importance of continuously aligning the latent representation with the textual narrative. By providing a global context, C2IA helps coordinate object arrangements, colors, and overall scene composition. Its absence causes certain scene-wide semantics to weaken, leading to lower scores.

\subsubsection{Effectiveness of the O2RA Module}
Analogously, omitting O2RA means only the textual prompt is aligned with the entire image, effectively removing any explicit bounding-box alignment. Table~\ref{tab:ablation} shows that this omission reduces image-level FID and CLIP by 0.4 and 0.0002, respectively, along with a region-level CLIP loss of 0.0006. Though these drops might seem modest compared to removing C2IA, they suggest that local constraints still play a pivotal role in guiding precise object placements. 
In fact, HCMA (w/o O2RA) remains superior to a model lacking both modules, which highlights that global semantics alone can provide partial gains. Yet, the absence of O2RA allows objects to deviate from their specified locations and attributes, limiting the method’s ability to faithfully represent complex or tightly constrained prompts.

\subsubsection{Combined Analysis}
Overall, these ablation results demonstrate that C2IA and O2RA fulfill complementary roles in HCMA. C2IA upholds a scene-wide semantic narrative by continuously matching the image representation to the text prompt, whereas O2RA refines individual bounding boxes to reflect their targeted content. Although each module independently boosts performance, the synergy is most evident when both are integrated: The complete HCMA model achieves the highest FID and CLIP scores at both the image and region levels. 
Notably, a slight trade-off emerges where global alignment can indirectly influence local generation, potentially affecting region-level fidelity, yet the net effect remains overwhelmingly positive. This interplay highlights the nuanced balance HCMA strikes between capturing the prompt’s overarching meaning and enforcing precise local constraints, ultimately yielding more coherent and spatially accurate generation.

\begin{figure*}
  \includegraphics[width=0.99\textwidth]{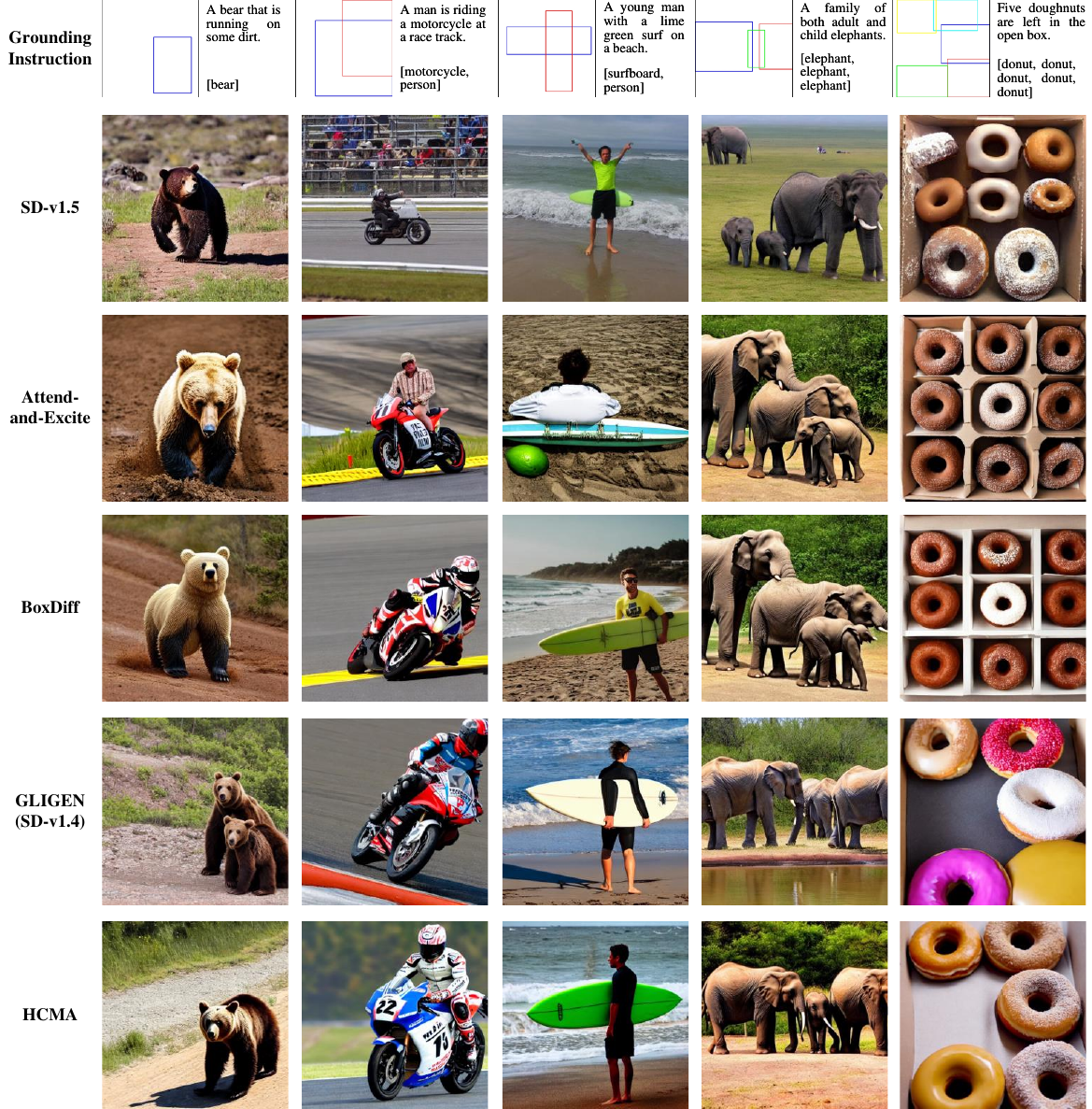}
  \caption{\small Visualization comparisons of different methods on the MS-COCO 2014 validation set.}
  \label{fig:qualitative_analysis}
\end{figure*}

\subsection{Qualitative Analysis}
Figure~\ref{fig:qualitative_analysis} shows a visual comparison between our HCMA framework and four baseline models, evaluated on five diverse samples that vary in complexity and object arrangements.

\noindent \textbf{(1) Single Object.}
In the first column, focusing on a simpler single-object scenario, Attend-and-Excite introduces artifacts, whereas SD-v1.5 generates a bear with unrealistic features. Although BoxDiff is a grounded text-to-image method, it disregards the specified layout constraints. GLIGEN (SD-v1.4) places a bear in the correct location but fails to respect the exact quantity indicated by both the text prompt and bounding box. In contrast, HCMA correctly positions a bear at the predefined spot, preserving a appearance (e.g., casting a plausible shadow) and matching the required object count.

\noindent \textbf{(2) Object Pair Relationships.}
The second and third columns depict a pair of objects that either interact or are arranged according to a prepositional relationship. Here, our HCMA approach precisely locates each distinct object in accordance with the bounding-box constraints and textual instructions, producing visually consistent scenes that exhibit minimal artifacts and maintain a lifelike quality. This outcome underscores HCMA’s capacity to handle notably more intricate spatial relationships beyond simple placements, revealing the synergy of its global and local alignment modules.

\noindent \textbf{(3) Counting and Classification.}
In the last two columns, HCMA not only generates the correct number of objects—such as multiple elephants in the fourth column and donuts in the fifth—but also distinguishes object categories (e.g., separating adult from child elephants). By contrast, GLIGEN (SD-v1.4) erroneously synthesizes three adult elephants in the fourth column and misclassifies baked items in the fifth column (producing four donuts plus an additional bread piece). These results emphasize HCMA’s proficiency in aligning textual instructions with both counting accuracy and fine-grained classification, while preserving high visual fidelity.

\section{Conclusion}
\label{sec:conclusion}
We have presented the \emph{Hierarchical Cross-Modal Alignment} (HCMA) framework, which unifies global text-image coherence with local bounding-box adherence for grounded text-to-image generation. By alternating between caption-to-image alignment and object-to-region alignment at each diffusion step, HCMA maintains high-level semantic fidelity while enforcing precise region-specific constraints. Experiments on the MS-COCO 2014 validation set show significant gains in FID and CLIP Score over state-of-the-art baselines, indicating that this layered alignment approach effectively addresses the shortcomings of methods focusing solely on global semantics or local placement. Looking ahead, HCMA could be extended by incorporating additional modalities (e.g., depth maps, reference images) to enrich layout control, or by using adaptive weighting between global and local modules to handle varying prompt complexities and domain-specific needs. Future work may also include human-in-the-loop feedback, domain adaptation for specialized datasets, or real-time interactive adjustments, further enhancing HCMA’s applicability to scenarios like virtual prototyping, creative media design, and immersive storytelling. Ultimately, HCMA offers a robust, flexible basis for tasks demanding both overarching textual fidelity and precise spatial configuration, paving the way for more advanced and controllable generative models.

%%
%% The next two lines define the bibliography style to be used, and
%% the bibliography file.
\bibliographystyle{ACM-Reference-Format}
\bibliography{ref}

\end{document}